\documentclass{article}
\usepackage{ijcai19}

\usepackage{times}
\usepackage{soul}
\usepackage{url}
\usepackage[hidelinks]{hyperref}
\usepackage[utf8]{inputenc}
\usepackage[small]{caption}
\usepackage{graphicx}
\usepackage{amsmath}
\usepackage{booktabs}
\usepackage{algorithm}
\usepackage{algorithmic}
\usepackage{tabularx}
\usepackage{amsmath,bm}
\usepackage{amssymb}
\usepackage{multirow}
\usepackage{subfigure}
\usepackage{mathrsfs}

\newtheorem{definition}{Definition}

\title{Clustering Uncertain Data via Representative Possible Worlds\\ with Consistency Learning}

\author{
Han Liu$^1$ \and
Xianchao Zhang$^2$ \and
Xiaotong Zhang$^1$ \and
Qimai Li$^1$ \and
Xiao-Ming Wu$^1$
\affiliations
$^1$The Hong Kong Polytechnic University, Hong Kong\\
$^2$Dalian University of Technology, China\\
\emails
liu.han.dut@gmail.com, xczhang@dlut.edu.cn, \\ zxt.dut@hotmail.com, \{csqmli,csxmwu\}@comp.polyu.edu.hk
}

\begin{document}

\maketitle

\begin{abstract}
Clustering uncertain data is an essential task in data mining for the internet of things. Possible world based algorithms seem promising for clustering uncertain data. However, there are two issues in existing possible world based algorithms: (1) They rely on all the possible worlds and treat them equally, but some marginal possible worlds may cause negative effects. (2) They do not well utilize the consistency among possible worlds, since they conduct clustering or construct the affinity matrix on each possible world independently. In this paper, we propose a representative possible world based consistent clustering (RPC) algorithm for uncertain data. First, by introducing representative loss and using Jensen-Shannon divergence as the distribution measure, we design a heuristic strategy for the selection of representative possible worlds, thus avoiding the negative effects caused by marginal possible worlds. Second, we integrate a consistency learning procedure into spectral clustering to deal with the representative possible worlds synergistically, thus utilizing the consistency to achieve better performance. Experimental results show that our proposed algorithm performs better than the state-of-the-art algorithms.
\end{abstract}

\section{Introduction}
Most existing clustering algorithms focus on certain data. However, due to various reasons like randomness in data generation and collection, imprecision in physical measurement, privacy concern and data staling \cite{DBLP:journals/tkde/AggarwalY09}, uncertain data is ubiquitous in many real applications, such as sensor networks, biomedical measurement, location tracking, meteorological forecasting and so on \cite{DBLP:journals/nn/ZhangLZ17}. Uncertain data has posed a serious challenge to existing clustering algorithms.

Several algorithms have been proposed for clustering uncertain data. Partition-based algorithms, e.g., UK-means \cite{DBLP:conf/pakdd/ChauCKN06} and UK-medoids \cite{DBLP:conf/sum/GulloPT08}, use expected distance or uncertain distance to extend traditional $k$-means or $k$-medoids to deal with uncertain data. However, they reduce complex probability distributions to a single probability distribution or a determinate value, thus cannot handle the uncertain information well \cite{DBLP:journals/nn/ZhangLZ17}. Density-based algorithms, e.g., FDBSCAN \cite{DBLP:conf/kdd/KriegelP05} and FOPTICS \cite{DBLP:conf/icdm/KriegelP05}, extend traditional DBSCAN \cite{DBLP:conf/kdd/EsterKSX96} or OPTICS \cite{DBLP:conf/sigmod/AnkerstBKS99} for clustering uncertain data by use of probabilistic definitions. However, they suffer from the unreasonable independent distance assumption \cite{DBLP:conf/kdd/ZufleESMZR14}, thus are difficult to obtain satisfactory performance.

Different from partition-based and density-based algorithms, possible world based algorithms, e.g., SC \cite{DBLP:conf/icde/VolkRHHL09} and REP \cite{DBLP:conf/kdd/ZufleESMZR14}, employ multiple independent and identically distributed realizations of an uncertain dataset to deal with data uncertainty, thus reducing the loss of uncertain information and avoiding the independent distance assumption. However, they have two unaddressed issues: (1) They rely on all the possible worlds and treat them equally, but some marginal possible worlds may cause negative effects on the clustering result. (2) They do not well utilize the consistency among possible worlds, since they conduct clustering or construct the affinity matrix on each possible world independently. Nevertheless, the consistency is important since different possible worlds can utilize it to transfer useful information for improving the performance.

In this paper, we propose a representative possible world based consistent clustering (RPC) algorithm for uncertain data, which improves existing algorithms from the following two aspects: (1) To alleviate the negative effects caused by marginal possible worlds, we introduce the definition of representative loss, use Jensen-Shannon divergence as the distribution measure, and then design a heuristic strategy for the selection of representative possible worlds. This strategy can be used by any possible world based algorithm to improve the performance. (2) To utilize the consistency to achieve better performance, we integrate a consistency learning procedure into spectral clustering to deal with the representative possible worlds synergistically. Extensive experimental results demonstrate the superiority of the proposed algorithm over the existing ones.

\section{Related Work}
\textbf{Traditional algorithms.} (1) \emph{Partition-based algorithms}: UK-means \cite{DBLP:conf/pakdd/ChauCKN06} is the first partition-based algorithm for clustering uncertain data. It extends the traditional $k$-means by using expected distance. To improve the efficiency of UK-means, \cite{DBLP:conf/icdm/KaoLCHC08,DBLP:journals/tkde/KaoLLCH10,DBLP:journals/is/NgaiKCCLCY11} use various pruning techniques to avoid the computation of redundant expected distances. CK-means \cite{DBLP:conf/icdm/LeeKC07} optimizes UK-means by resorting to the moment of inertia of rigid bodies. DUK-means \cite{DBLP:journals/tnn/ZhouCCWL18} is an improved version of UK-means, which is specifically designed for distributed network environment. UK-medoids \cite{DBLP:conf/sum/GulloPT08} employs uncertain distance to extend the traditional $k$-medoids. MMVar \cite{DBLP:conf/icdm/GulloPT10} uses a novel objective function which aims to minimize the variance of cluster mixture models. UCPC \cite{DBLP:journals/pvldb/GulloT12} introduces the notion of uncertain centroid and it is a local search based heuristic algorithm. All these algorithms can deal with uncertain data to some extent. However, they reduce complex probability distributions to a single probability distribution or a determinate value, thus cannot handle the uncertain information well \cite{DBLP:journals/nn/ZhangLZ17}. (2) \emph{Density-based algorithms}: FDBSCAN \cite{DBLP:conf/kdd/KriegelP05} and FOPTICS \cite{DBLP:conf/icdm/KriegelP05} are the first density-based and hierarchical density-based algorithms for clustering uncertain data respectively. They introduce a series of probabilistic definitions to extend the traditional DBSCAN or OPTICS. Zhang et al. \cite{zhang2014novel,DBLP:journals/nn/ZhangLZ17} analyze the limitations in FDBSCAN and FOPTICS, and propose a novel density-based algorithm PDBSCAN for clustering uncertain data. However, these algorithms rely on the unreasonable independent distance assumption \cite{DBLP:conf/kdd/ZufleESMZR14}, thus are difficult to obtain satisfactory clustering results.

\noindent\textbf{Possible world based algorithms.} SC \cite{DBLP:conf/icde/VolkRHHL09} is the first possible world based algorithm for clustering uncertain data. It conducts clustering on each possible world independently and integrates the clustering results into one final result. REP \cite{DBLP:conf/kdd/ZufleESMZR14} also conducts clustering on each possible world independently, but it selects the representative clustering result as the final result. Recently, \cite{liu2018possible} tries to leverage the consistency principle for clustering uncertain data. It constructs the affinity matrix for each possible world independently and then learns a consensus affinity matrix for clustering uncertain data. However, the consistency learning method introduced in \cite{liu2018possible} lacks the procedure of updating the affinity matrix of each possible world, thus reducing the ability of consistency learning. Possible world based algorithms avoid the issues in traditional algorithms and seem more promising for clustering uncertain data. However, as we point out hereinafter, there are some unaddressed issues in existing possible world based algorithms.

\section{Preliminaries}
\subsection{Consistency Principle}
Consistency principle is a common assumption, which has been widely used in many machine learning methods \cite{DBLP:conf/aaai/LiuL0DYZ17}. Its definition is as follows \cite{DBLP:conf/icml/WangZ10}.
\begin{definition}
(Consistency principle). Given a dataset which has multiple representations, consistency principle refers to an assumption that the class labels and cluster structures of the multiple representations are consistent.
\end{definition}

By using consistency principle to minimize the disagreement of different representations, we can improve the algorithm performance greatly. The detailed proof can refer to \cite{DBLP:conf/nips/DasguptaLM01}.

\subsection{Uncertain Data and Possible World}
Uncertain data can be considered at table, tuple or attribute level \cite{DBLP:journals/vldb/SarmaBHNW09}. For uncertain data clustering, we mainly focus on attribute level uncertainty. That is to say, each uncertain object is represented as a random variable with a probability distribution, which is associated with the probability that the object appears at any position in a multidimensional space.

Possible world is an effective tool to model uncertain data \cite{DBLP:conf/pods/DalviS07,DBLP:journals/vldb/SarmaBHNW09}. Its definition is as follows \cite{DBLP:series/ads/HuaP11}.
\begin{definition}
(Possible world). Let $UD = \left\{ {{O_1},{O_2},...,{O_n}} \right\}$ be an uncertain dataset. A possible world $pw = \left\{ {{o_1},{o_2},...,{o_n}} \right\} (o_i\in O_i)$ is a set of instances such that each instance is taken from each corresponding uncertain object. Let $PW$ be the set of all the possible worlds, $P(pw)$ be the existence probability of $pw$, then $\sum_{pw\in PW} P(pw)=1$.
\end{definition}

Possible world can be generated through the inversion method. Due to space limitation, more information and proofs can refer to \cite{Devroye86,jampani2008mcdb}.

\subsubsection{Consistency Principle for Possible World}

\begin{figure}[t]
  \centering
    \includegraphics[height=0.20\columnwidth, width=0.85\columnwidth]{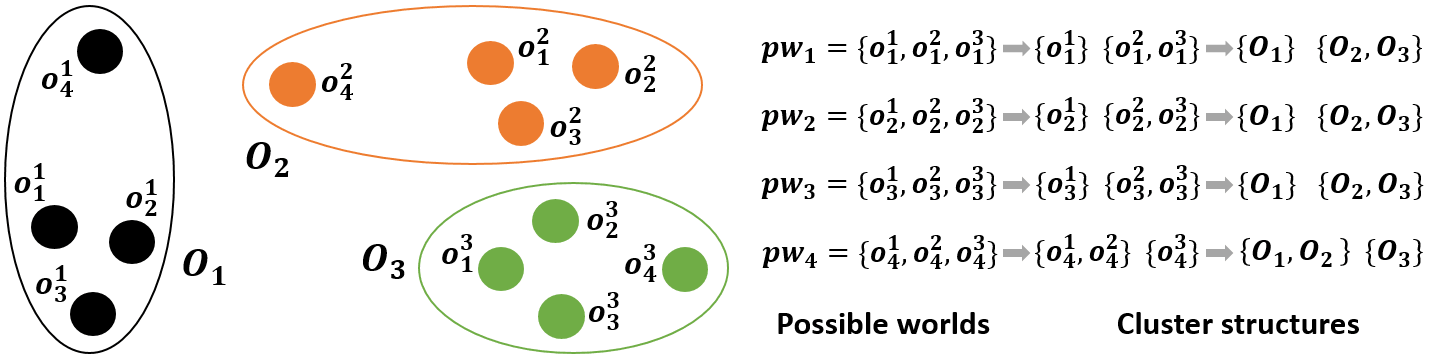}
  \caption{Consistency principle for possible world.}
  \label{figure1}
\end{figure}

According to the definition of possible world, different possible worlds come from the same uncertain dataset and they are a number of independent and identically distributed realizations of an uncertain dataset \cite{DBLP:series/ads/HuaP11}. Therefore, if we treat each possible world as one representation of the uncertain dataset, by the concept of consistency principle, we can have the following consistency principle for possible world: \emph{the class labels and cluster structures of different possible worlds are consistent}.

In general, the consistency principle for possible world conforms to the reality well, i.e., in most cases the class labels and cluster structures of different possible worlds are consistent. For example, in Figure 1, $O_1$, $O_2$, $O_3$ are uncertain objects, and $o^i_1$, $o^i_2$, $o^i_3$, $o^i_4$ are the possible instances of $O_i$ ($i \in \{1,2,3\}$). If we divide $O_1$, $O_2$, $O_3$ into two clusters, based on the geometric information, $O_1$ should belong to one cluster, $O_2$ and $O_3$ should belong to the other cluster. For the possible worlds $pw_1$, $pw_2$, $pw_3$ with their components shown in Figure 1, it is easy to find that their class labels and cluster structures are consistent.

However, the consistency principle for possible world is not absolute. In some cases, abnormal possible worlds violate the principle, and we call this kind of possible worlds as the marginal ones. Formally, the definition of marginal possible world is as follows.
\begin{definition}
(Marginal possible world). Let $PW$ be the set of all the possible worlds, marginal possible world refers to the possible world whose class label and cluster structure have large differences with most possible worlds in $PW$.
\end{definition}

For example, in Figure 1, $pw_4$ is a possible world which consists of some abnormal instances. As the class label and cluster structure of $pw_4$ are very different from most possible worlds, $pw_4$ is a marginal possible world.

\subsection{Unaddressed Issues}

\noindent \textbf{(1) Negative effects caused by marginal possible worlds.} Existing possible world based algorithms rely on all the possible worlds and treat them equally. However, marginal possible worlds belong to the abnormal ones, their class labels and cluster structures have large differences with most possible worlds, which may disturb the integrating or selecting procedure of existing possible world based algorithms and cause negative effects on the clustering result. To solve this issue, we propose to select some representative possible worlds to filter out marginal possible worlds. By \textbf{representative possible worlds} we mean a subset of all the possible worlds which has a strong ability to represent all the possible worlds. As marginal possible worlds are abnormal and their representative ability is weak, we can filter out marginal possible worlds and avoid the negative effects by selecting representative possible worlds.

\noindent \textbf{(2) Utilizing the consistency principle for possible world not well.} The consistency principle makes it possible to transfer useful information among different possible worlds, which can potentially improve the clustering quality. However, existing possible world based algorithms conduct clustering or construct the affinity matrix on each possible world independently, thus cannot well utilize the consistency among possible worlds. To tackle this issue, we propose a consistent spectral clustering method which can update the eigenvector matrix of each possible world iteratively and minimize the disagreement of different possible worlds, thus better achieving the consistency learning and improving the performance.

\section{The Proposed Algorithm}
The proposed algorithm consists of two parts: selecting representative possible worlds and consistent spectral clustering.

Given an uncertain dataset $UD=\{O_1,O_2,...,O_n\}$ in a $d$-dimensional independent space, $PW=\{pw_i|i=1,2,...,M\}$, $PWR=\{pwr_j|j=1,2,...,R\}$, $PWU=\{pwu_k|k=1,2,...,M-R\}$ respectively denote the set of all the possible worlds, the representative possible world set and the unrepresentative possible world set. $M$, $R$ and $M-R$ respectively denote the number of possible worlds in $PW$, $PWR$ and $PWU$. Here $PW=PWR\cup PWU$.

\subsection{Selecting Representative Possible Worlds}
By selecting representative possible worlds, we can filter out marginal possible worlds and avoid the waste of time caused by redundant possible worlds. In order to select representative possible worlds, we introduce the definition of representative loss, use Jensen-Shannon divergence as the distribution measure, and then design a heuristic strategy for the selection of representative possible worlds.

\subsubsection{Representative Loss}
Intuitively, given any two possible worlds $pw$ and $pw'$, if we want to use $pw$ to represent $pw'$, then the smaller the difference between $pw$ and $pw'$, the less the loss that $pw$ represents $pw'$. We aim to select $PWR$ from $PW$ to represent $PW$. As $PW=PWR\cup PWU$ and the loss that $PWR$ represents $PWR$ is equal to $0$, then the loss that $PWR$ represents $PW$ is equal to the loss that $PWR$ represents $PWU$. Based on these observations, we have the following definition.
\begin{definition}
(Representative loss). Let $PWR$ be the representative possible world set and $pwr_j \in PWR$, $PWU$ be the unrepresentative possible world set and $pwu_k \in PWU$. If using $PWR$ to represent $PWU$, then the representative loss, denoted by $L(PWR\rightarrow PWU)$, can be defined as:
\begin{equation}
L(PWR\rightarrow PWU)= \sum\limits_{k=1}^{M-R}\mathop {\min }\limits_{pwr_j}\Phi(pwr_j,pwu_k),
\label{eq1}
\end{equation}
\end{definition}
where $\Phi(pwr_j,pwu_k)$ is the difference between $pwr_j$ and $pwu_k$, $M-R$ is the number of possible worlds in $PWU$.

From this definition, it can be seen that if we know how to compute the difference between possible worlds, we can get the representative loss that $PWR$ represents $PWU$, i.e., the representative loss that $PWR$ represents $PW$.

\subsubsection{Jensen-Shannon Divergence between Possible Worlds}
As a possible world can be regarded as a probability distribution, we can compute the difference between possible worlds by Jensen-Shannon divergence \cite{lin1991divergence}. Compared with KL divergence \cite{kullback1951information}, Jensen-Shannon divergence is symmetric and finite, therefore it is more suitable as the representative loss measure.

Given any two possible worlds $pw$ and $pw'$, the Jensen-Shannon divergence between them can be defined as:
\begin{equation}
\begin{split}
JSD(pw||pw')=\frac{1}{2}D(P_{pw}||H)+\frac{1}{2}D(P_{pw'}||H),
\label{eq2}
\end{split}
\end{equation}
where $P_{pw}$ and $P_{pw'}$ are the probability distributions of $pw$ and $pw'$ respectively, and $H=\frac{1}{2}(P_{pw}+P_{pw'})$. $D(P||Q)$ is the KL divergence between two probability distributions $P$ and $Q$. For continuous probability distributions $P$ and $Q$ with a variable $x$ in a domain $\mathbb{D}$, $D(P||Q)$ is defined as:
\begin{equation}
D(P||Q)=\int_\mathbb{D}f(x)log\frac{f(x)}{g(x)}dx,
\label{eq3}
\end{equation}
where $f(x)$ and $g(x)$ are the probability density functions of $P$ and $Q$. $D(P||Q)$ can also be expressed as:
\begin{equation}
D(P||Q)=E(log\frac{f(x)}{g(x)}),
\label{eq4}
\end{equation}
where $E$ denotes the expectation. According to the law of large numbers and Eq.(\ref{eq4}), given a sample set $S$, $D(P||Q)$ can be estimated by:
\begin{equation}
D(P||Q)=\frac{1}{|S|}\sum \limits_{x\in S} log \frac{f(x)}{g(x)},
\label{eq5}
\end{equation}
where $|S|$ denotes the number of objects in $S$.

We employ the kernel density estimation method \cite{silverman1986density} to obtain the probability density functions $f_{pw}$ and $f_{pw'}$ of the probability distributions $P_{pw}$ and $P_{pw'}$. Specifically, $f_{pw}$ can be estimated as:
\begin{equation}
f_{pw}(x) = \frac{1}
{{ |pw| \prod\nolimits_{j = 1}^d {h_j } }}\sum\limits_{o \in pw} {\prod\limits_{j = 1}^d {K(\frac{{x.D_j  - o.D_j }}
{{h_j }})} }.
\label{eq6}
\end{equation}

In Eq.(\ref{eq6}), $o$ denotes an object in $pw$ and it can be represented by $(o.D_1, o.D_2, ..., o.D_d)$, $d$ denotes the total dimensionality, and $|pw|$ denotes the number of objects in $pw$. $K$ denotes the kernel function, and we use the most common Gaussian kernel function. $h_j$ denotes the bandwidth of the $j$-th dimension. For Gaussian kernel function, we set $h_j=1.06\times\delta_j|pw|^{-\frac{1}{5}}$ according to the Silverman's rule of thumb \cite{silverman1986density}, where $\delta_j$ is the standard deviation of the $j$-th dimension of the objects in $pw$.

By using Jensen-Shannon divergence as the distribution measure to compute the difference between possible worlds, i.e., replacing $\Phi(pwr_j,pwu_k)$ in Eq.(\ref{eq1}) with $JSD(pwr_j||pwu_k)$, we can get the representative loss.

\subsubsection{Selection Strategy}
Our goal is to select a given number of possible worlds as the representative possible worlds. In general, a good representative possible world set should have a strong representative ability, i.e., its corresponding representative loss should be small. Inspired by this observation, we propose the following selection strategy:

\emph{Let $PWR$ be the representative possible world set, and $PWU$ be the unrepresentative possible world set. Now select a possible world $pwu^*$ from $PWU$ and move $pwu^*$ to $PWR$, if we want the new representative possible world set $PWR\cup pwu^*$ to be the best, then the selection strategy should ensure the representative loss that $PWR\cup pwu^*$ represents $PWU\backslash pwu^*$ to be the minimum. Formally:}
\begin{equation}
pwu^*=\mathop{\arg\min}\limits_{pwu^*}L(PWR\cup pwu^* \rightarrow PWU\backslash pwu^*).
\label{eq7}
\end{equation}

From Eq.(\ref{eq7}), it can be seen that $pwu^*$ should have a strong representative ability. Marginal possible worlds are the abnormal ones and their representative ability is poor, therefore this selection strategy can filter out marginal possible worlds.

Based on the selection strategy, we design a heuristic method to select the representative possible worlds, which is shown in Algorithm 1 (Part 1).

\begin{algorithm}[t]
\linespread{1.35}
\tiny
   \caption{RPC}
\begin{algorithmic}
\REQUIRE Uncertain dataset $UD=\{O_1,O_2,...,O_n\}$, the number of clusters $k$, the number \\~~~~~~~~ of all the possible worlds $M$, the number of representative possible worlds $R$
\ENSURE The clusters $C_1,C_2,...,C_k$
   \STATE ~~~~~~~\textbf{Part 1: Selecting representative possible worlds (Lines 1-5)}
   \STATE 1:~~~~Generate $PW$, initialize $PWR= \emptyset$ and $PWU = PW$, and calculate the $JSD$ \\~~~~~~ between any two possible worlds in $PW$
   \STATE 2:~~~~\textbf{Repeat}
   \STATE 3:~~~~~~~~Select a possible world $pwu^*$ from $PWU$ by Eq.(\ref{eq7})
   \STATE 4:~~~~~~~~$PWR \leftarrow PWR \cup pwu^*$, $PWU \leftarrow PWU \backslash pwu^*$
   \STATE 5:~~~~\textbf{Until} $|PWR|\geqslant R$, $|PWR|$ denotes the current number of possible worlds in $PWR$
   \STATE ~~~~~~~\textbf{Part 2: Consistent spectral clustering (Lines 6-12)}
   \STATE 6:~~~~For $\forall pwr_j \in PWR$, compute $W^{(j)}$, $D^{(j)}$, $L^{(j)}$
   \STATE 7:~~~~For $\forall pwr_j \in PWR$, compute the $k$ eigenvectors corresponding to the $k$ largest \\~~~~~~ eigenvalues of $L^{(j)}$ and use them to initialize the corresponding $U^{(j)}$
   \STATE 8:~~~~\textbf{Repeat}
   \STATE 9:~~~~~~~~Update $U^*$ by solving Eq.(\ref{eq13})
   \STATE 10:~~~~~~Update each $U^{(j)}$ by solving Eq.(\ref{eq15})
   \STATE 11:~~~\textbf{Until} Eq.(\ref{eq11}) is convergent
   \STATE 12:~~~Run $k$-means on $U^*$ and get the clusters $C_1,C_2,...,C_k$
\end{algorithmic}
\end{algorithm}

\subsection{Consistent Spectral Clustering}
We integrate a consistency learning procedure into spectral clustering to deal with the representative possible worlds synergistically.

\subsubsection{Spectral Clustering}
Assume that $pwr_j$ is a possible world from the representative possible world set $PWR$ and $PWR=\{pwr_j|j=1,2,...,R\}$, where $R$ denotes the number of possible worlds in $PWR$. $W^{(j)}$ is the similarity matrix of $pwr_j$, which is computed by the Gaussian kernel. $L^{(j)}$ is the normalized Laplacian matrix of $pwr_j$ and $L^{(j)}=D^{(j)^{-\frac{1}{2}}}W^{(j)}D^{(j)^{-\frac{1}{2}}}$. $D^{(j)}$ is a diagonal matrix and $D^{(j)}(i,i)=\sum \limits_{l=1}^{n} W^{(j)}(i,l)$, where $n$ denotes the number of objects in $pwr_j$. For the representative possible world $pwr_j$, the objective function of spectral clustering is:
\begin{equation}
\max \limits_{U^{(j)}} tr(U^{(j)^T}L^{(j)}U^{(j)}), ~~s.t.~~U^{(j)^T}U^{(j)}=I,
\label{eq8}
\end{equation}
where $tr(\cdot)$ denotes the trace of a matrix, and the solution of $U^{(j)} \in \mathbb{R}^{n\times k}$ is composed by $k$ eigenvectors corresponding to the $k$ largest eigenvalues of $L^{(j)}$.

\subsubsection{Consistency Learning}
The eigenvector matrix $U^{(j)}$ can reflect the cluster structure of the representative possible world $pwr_j$. To meet the requirement of consistency, we assume that each eigenvector matrix $U^{(j)} \in \mathbb{R}^{n\times k}$ tends to a common eigenvector matrix $U^* \in \mathbb{R}^{n\times k}$. Then by minimizing the disagreement between each $U^{(j)}$ and $U^*$, we can achieve the consistency learning among different possible worlds. For the disagreement between $U^{(j)}$ and $U^*$, we use the squared Euclidean distance between the similarity matrices to measure it:
\begin{equation}
\begin{split}
Dis(U^{(j)}, U^*)=||S_{U^{(j)}}-S_{ U^*}||^2_F, \\ s.t. ~~U^{(j)^T}U^{(j)}=I,~~U^{*^T}U^*=I,~
\end{split}
\label{eq9}
\end{equation}
where $S_{U^{(j)}}$ and $S_{ U^*}$ denote the similarity matrices of $U^{(j)}$ and $U^*$, and $||\cdot||_F$ denotes the Frobenius norm of the matrix.

Considering the feasibility of optimization, we use the commonly adopted inner product to compute the similarity matrix, i.e., $S_{U^{(j)}}=U^{(j)}U^{(j)^T}$. Then with some manipulations, minimizing Eq.(\ref{eq9}) can be transformed as:
\begin{equation}
\begin{split}
\max \limits_{U^{(j)}, U^*}tr(U^{(j)}U^{(j)^T}U^*U^{*^T}),~~~~~ \\ s.t. ~~U^{(j)^T}U^{(j)}=I,~~U^{*^T}U^*=I.~~
\end{split}
\label{eq10}
\end{equation}

\subsubsection{Overall Objective Function and Optimization}
By integrating the objective functions of spectral clustering and consistency learning, we can get the overall objective function of consistent spectral clustering as follows:
\begin{equation}
\begin{split}
\max \limits_{U^{(j)}, U^*} \sum \limits_{j=1}^R (tr(U^{(j)^T}L^{(j)}U^{(j)})+  tr(U^{(j)}U^{(j)^T}U^*U^{*^T})),\\
~~s.t.~~ U^{(j)^T}U^{(j)}=I, ~~1\leqslant j\leqslant R, ~~U^{*^T}U^*=I.~~~~~~~~~
\end{split}
\label{eq11}
\end{equation}

For Eq.(\ref{eq11}), we can employ the alternative iteration method to solve it.

(1) Optimizing Eq.(\ref{eq11}) with respect to $U^*$. Fix each $U^{(j)}$, then Eq.(\ref{eq11}) becomes:
\begin{equation}
\begin{split}
\max \limits_{U^*} \sum \limits_{j=1}^R tr(U^{(j)}U^{(j)^T}U^*U^{*^T}),
~~s.t. ~~U^{*^T}U^*=I.
\end{split}
\label{eq12}
\end{equation}

Eq.(\ref{eq12}) can be written as:
\begin{equation}
\begin{split}
\max \limits_{U^*} tr (U^{*^T}(\sum \limits_{j=1}^R  U^{(j)}U^{(j)^T})U^*),
~s.t. ~U^{*^T}U^*=I.
\end{split}
\label{eq13}
\end{equation}

It is easy to find that optimizing Eq.(\ref{eq13}) is equivalent to solve the standard spectral clustering with a modified Laplacian matrix $\sum \limits_{j=1}^R U^{(j)}U^{(j)^T}$, i.e., the solution of $U^*$ is composed by $k$ eigenvectors corresponding to the $k$ largest eigenvalues of $\sum \limits_{j=1}^R U^{(j)}U^{(j)^T}$.

(2) Optimizing Eq.(\ref{eq11}) with respect to one of the $U^{(j)}$s. Fix the other $U^{(j)}$s and $U^*$, then Eq.(\ref{eq11}) becomes:
\begin{equation}
\begin{split}
\max \limits_{U^{(j)}} tr(U^{(j)^T}L^{(j)}U^{(j)})+ tr(U^{(j)}U^{(j)^T}U^*U^{*^T}),\\
s.t.~~ U^{(j)^T}U^{(j)}=I. ~~~~~~~~~~~~~~~~~~~~~~~
\end{split}
\label{eq14}
\end{equation}

Eq.(\ref{eq14}) can be written as:
\begin{equation}
\begin{split}
\max \limits_{U^{(j)}} tr(U^{(j)^T}(L^{(j)}+U^*U^{*^T}) U^{(j)}),\\
s.t.~~ U^{(j)^T}U^{(j)}=I. ~~~~~~~~~~~~~~~~
\end{split}
\label{eq15}
\end{equation}

Optimizing Eq.(\ref{eq15}) is similar with optimizing Eq.(\ref{eq13}), therefore the solution of $U^{(j)}$ is composed by $k$ eigenvectors corresponding to the $k$ largest eigenvalues of $L^{(j)}+U^*U^{*^T}$.

The overall procedure of consistent spectral clustering is shown in Algorithm 1 (Part 2).

\section{Experiments}
\subsection{Datasets}
\textbf{Real benchmark datasets.} We conduct experiments on 6 real benchmark datasets. The details of the datasets are shown in Table 1. These datasets are originally established as collections of data with determinate values, we follow the method in \cite{DBLP:conf/kdd/ZufleESMZR14,DBLP:journals/nn/ZhangLZ17} to generate the uncertainty of Gaussian distribution for these datasets.

\begin{table}[t]
\scriptsize
\caption{Real benchmark datasets.}
\vskip -0.1in
\label{dataset-table}
\begin{center}
\begin{tabular}{llll}
\midrule
Dataset & \#Objects & \#Attributes & \#Classes \\
\midrule
Wine    & 178 & 13 & 3\\
Ecoli   & 327 & 7 & 5\\
Image   & 2310 & 19 & 7\\
Libras   & 360 & 90 & 15\\
USPS   & 929 & 256 & 10\\
Waveform   & 5000 & 21 & 3\\
\midrule
\end{tabular}
\end{center}
\vskip -0.1in
\end{table}

\noindent\textbf{Real world uncertain datasets.} We also use 3 real world uncertain datasets: Movement (http://archive.ics.uci.edu/ml/), NBA (http://espn.go.com/nba/) and Weather (http://bcc.ncc-cma.net/) to perform experiments.

(1) Movement: it consists of 13197 radio signal records about 314 temporal sequences from a wireless sensor network deployed in real-world office environments. Each record has four dimensions which are respectively corresponding to four sensor nodes. According to user movement path, the dataset is divided into six classes. Each temporal sequence is treated as an uncertain object and each record of the temporal sequence is treated as a possible value of the uncertain object.

(2) NBA: it consists of 2197 records about the top 300 players in ESPN 2015 rank. Each record has five dimensions: points, rebounds, assists, steals and blocks. According to season average performance, they are divided into three classes: star/key/role player. Each player is treated as an uncertain object and each season average performance of the player is treated as a possible value of the uncertain object.

(3) Weather: it consists of 18360 records about 153 weather stations around China. Each station contains the monthly average weather condition from 2006 to 2015. Each record has two dimensions: average temperature and average precipitation. Each station is labeled with a climate type. We have three types of climates: temperate continental climate, temperate monsoon climate and tropical/subtropical monsoon climate. The stations with the same label are considered to be in the same class. Each station is treated as an uncertain object and each monthly average weather condition of the station is treated as a possible value of the uncertain object.

\begin{table*}[t]
\scriptsize
\caption{Clustering results in terms of effectiveness.}
\vskip -0.1in
\label{result-table1}
\begin{center}
\begin{tabular}{ccccccccccccccc}
\midrule
Dataset  & Metric & UKM & CKM & UKMD & MMV & UCPC& FDB & FOP & PDB &SC & REP &RP-SC & RP-REP& RPC \\
\midrule
\multirow{2}{*}{Wine}
&ACC  &0.8343 & 0.8213 & 0.8163 & 0.8056 & 0.8444 & 0.7247 & 0.7528 & 0.7303 & 0.7079 & 0.7416 & 0.7360 & 0.8034 & \textbf{0.9663} \\
&NMI  &0.7091 & 0.6435 & 0.6880 & 0.6419 & 0.6795 & 0.5562 & 0.6277 & 0.6195 & 0.4817 & 0.5460 & 0.5434 & 0.5823 & \textbf{0.8782} \\
\multirow{2}{*}{Ecoli}
&ACC  &0.6321 & 0.6300 & 0.6352 & 0.6309 & 0.5765 & 0.5260 & 0.6667 & 0.6575 & 0.6728 & 0.6667 & 0.7034 & 0.7278 & \textbf{0.8055} \\
&NMI  &0.6102 & 0.6362 & 0.5912 & 0.5569 & 0.5588 & 0.2040 & 0.5917 & 0.5536 & 0.4973 & 0.5124 & 0.5858 & 0.5860 & \textbf{0.6871} \\
\multirow{2}{*}{Image}
&ACC  &0.6639 & 0.6425 & 0.6980 & 0.5945 & 0.5819 & 0.5494 & 0.7177 & 0.7299 & 0.5870 & 0.5636 & 0.6576 & 0.6545 & \textbf{0.8350} \\
&NMI  &0.7115 & 0.6601 & 0.6818 & 0.6070 & 0.5933 & 0.6849 & 0.7464 & 0.7647 & 0.6182 & 0.5871 & 0.6925 & 0.6854 & \textbf{0.7838} \\
\multirow{2}{*}{Libras}
&ACC  &0.5322 & 0.5053 & 0.5294 & 0.4211 & 0.4414 & 0.2528 & 0.3417 & 0.3222 & 0.2611 & 0.3167 & 0.3083 & 0.3778 & \textbf{0.6006} \\
&NMI  &0.6583 & 0.6555 & 0.6314 & 0.5490 & 0.5752 & 0.4814 & 0.5742 & 0.5637 & 0.4997 & 0.5752 & 0.5292 & 0.6100 & \textbf{0.7056} \\
\multirow{2}{*}{USPS}
&ACC  &0.6220 & 0.6245 & 0.6499 & 0.5107 & 0.5269 & 0.4101 & 0.4769 & 0.4833 & 0.4101 & 0.4456 & 0.4327 & 0.4639 & \textbf{0.7658} \\
&NMI  &0.6797 & 0.6539 & 0.6574 & 0.5338 & 0.5326 & 0.4741 & 0.5622 & 0.5782 & 0.4939 & 0.5386 & 0.5242 & 0.5639 & \textbf{0.8082} \\
\multirow{2}{*}{Waveform}
&ACC  &0.8381 & 0.8382 & 0.7080 & 0.6569 & 0.6542 & 0.3428 & 0.3294 & 0.5938 & 0.4366 & 0.4386 & 0.4956 & 0.4524 & \textbf{0.9618} \\
&NMI  &0.6667 & 0.6697 & 0.4554 & 0.4397 & 0.4046 & 0.0602 & 0.0667 & 0.2975 & 0.0931 & 0.0489 & 0.1061 & 0.1465 & \textbf{0.8400} \\
\midrule
\multirow{2}{*}{Movement}
&ACC  &0.3490 & 0.3341 & 0.3478 & 0.3427 & 0.3494 & 0.2834 & 0.2643 & 0.3121 & 0.2548 & 0.2866 & 0.2866 & 0.3153 & \textbf{0.4315} \\
&NMI  &0.2133 & 0.1935 & 0.2172 & 0.1837 & 0.1985 & 0.0445 & 0.0791 & 0.1170 & 0.0688 & 0.0975 & 0.1350 & 0.1361 & \textbf{0.2584} \\
\multirow{2}{*}{NBA}
&ACC  &0.5463 & 0.5457 & 0.5403 & 0.5257 & 0.5473 & 0.5667 & 0.5067 & 0.5867 & 0.5133 & 0.5433 & 0.5667 & 0.5700 & \textbf{0.6133} \\
&NMI  &0.1591 & 0.1648 & 0.1558 & 0.1647 & 0.1690 & 0.1443 & 0.0918 & 0.1759 & 0.0671 & 0.1446 & 0.1563 & 0.1571 & \textbf{0.1919} \\
\multirow{2}{*}{Weather}
&ACC  &0.5869 & 0.6144 & 0.5961 & 0.6105 & 0.6033 & 0.5882 & 0.5163 & 0.6993 & 0.5294 & 0.5490 & 0.6340 & 0.6405 & \textbf{0.7176} \\
&NMI  &0.4892 & 0.4690 & 0.4486 & 0.4183 & 0.3747 & 0.1937 & 0.4575 & 0.5277 & 0.2714 & 0.2761 & 0.3133 & 0.3724 & \textbf{0.5842} \\
\midrule
\end{tabular}
\end{center}
\vskip -0.15in
\end{table*}

\subsection{Experimental Setup}
\textbf{Baselines.} We compare RPC with the state-of-the-art clustering algorithms for uncertain data, including UK-means (UKM), CK-means (CKM), UK-medoids (UKMD), MMVar (MMV), UCPC, FDBSCAN (FDB), FOPTICS (FOP), PDBSCAN (PDB), SC and REP. We also compare with the improved versions of SC and REP, which use our proposed selection strategy to select the representative possible worlds and then perform the original SC and REP on the representative possible worlds, and we call them RP-SC and RP-REP.

\noindent \textbf{Settings.} For UK-means, CK-means, UK-medoids, MMVar, UCPC and RPC, the sets of initial centroids or partitions are randomly selected. To avoid that the clustering results are affected by random chance, we average the results over 10 different runs. For FDBSCAN, FOPTICS, PDBSCAN, SC, REP, RP-SC and RP-REP, since these algorithms are sensitive to parameters, we adjust the parameters continuously until the performance of each method becomes the best and stable. The methods of determining the parameters can refer to \cite{DBLP:conf/kdd/KriegelP05,DBLP:conf/icdm/KriegelP05,DBLP:journals/nn/ZhangLZ17,DBLP:conf/icde/VolkRHHL09,DBLP:conf/kdd/ZufleESMZR14}.

\noindent \textbf{Evaluation metrics.} We adopt two widely used evaluation metrics \cite{DBLP:books/daglib/0021593}: clustering accuracy (ACC) and normalized mutual information (NMI) to evaluate the clustering performance.

\subsection{Parameter Investigation for RPC}

\begin{figure}[t]
  \centering
  \subfigure[Movement]{
    \includegraphics[height=0.23\columnwidth, width=0.29\columnwidth]{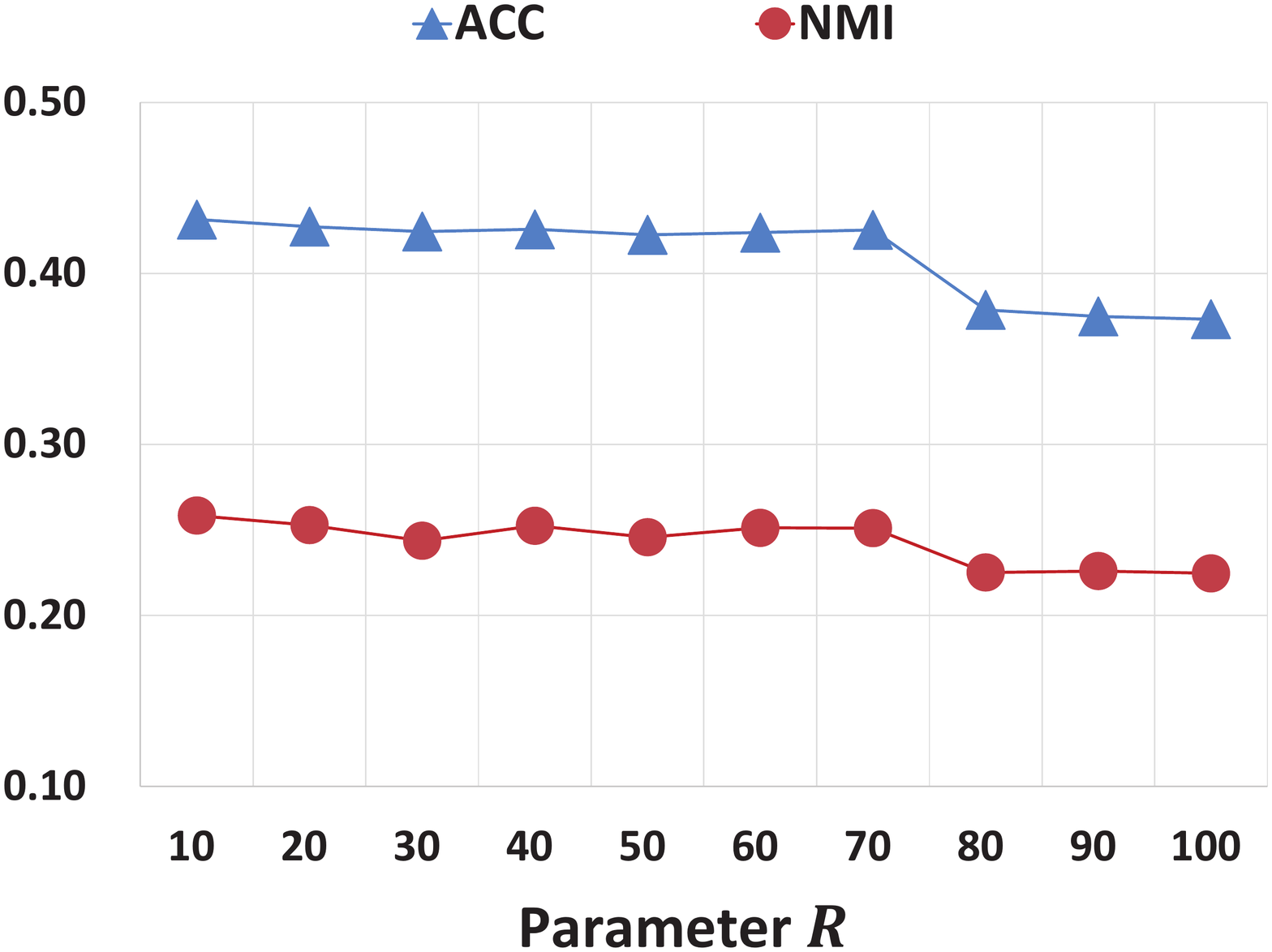}}
  \subfigure[NBA]{
    \includegraphics[height=0.23\columnwidth, width=0.29\columnwidth]{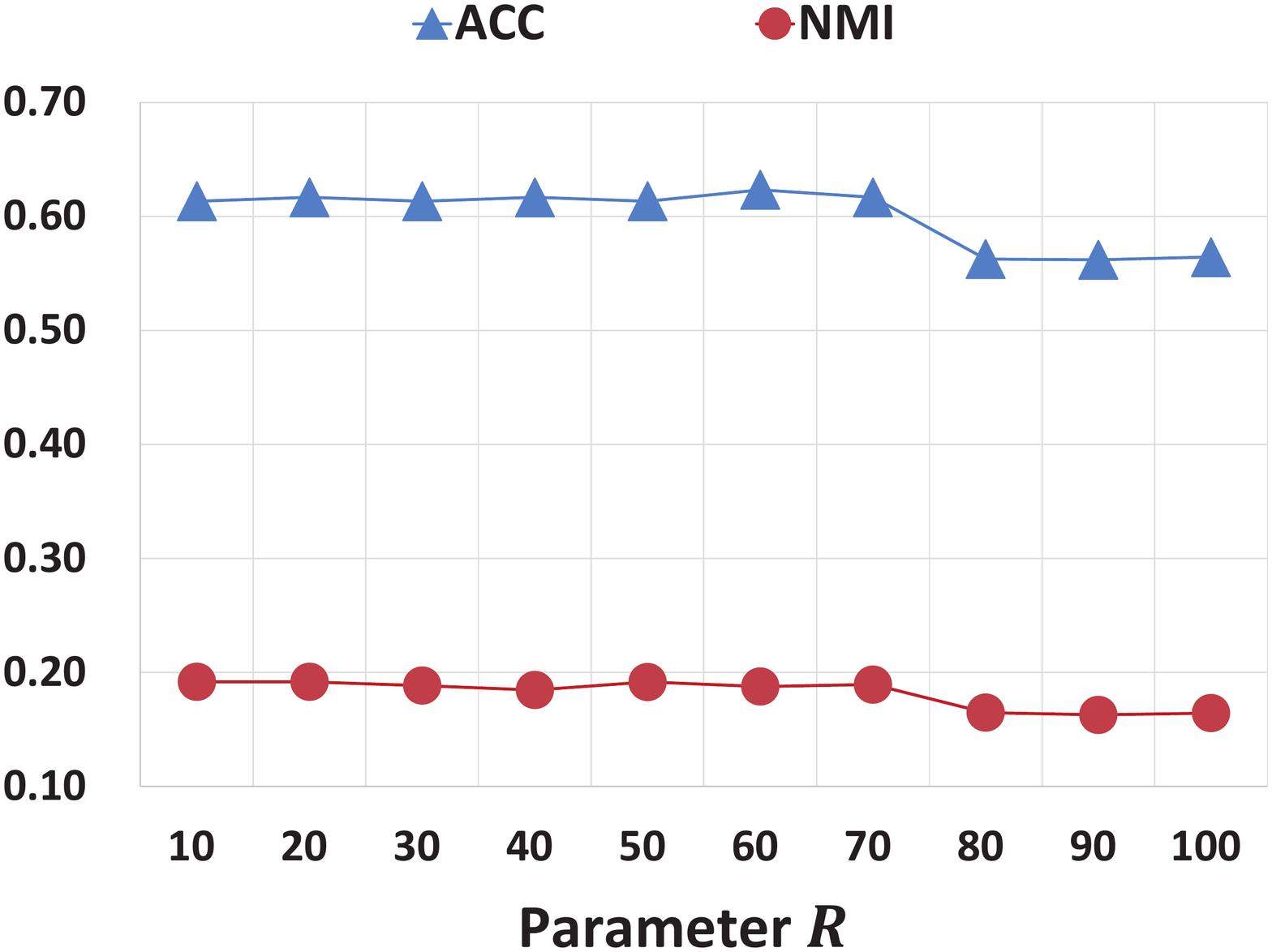}}
  \subfigure[Weather]{
    \includegraphics[height=0.23\columnwidth, width=0.29\columnwidth]{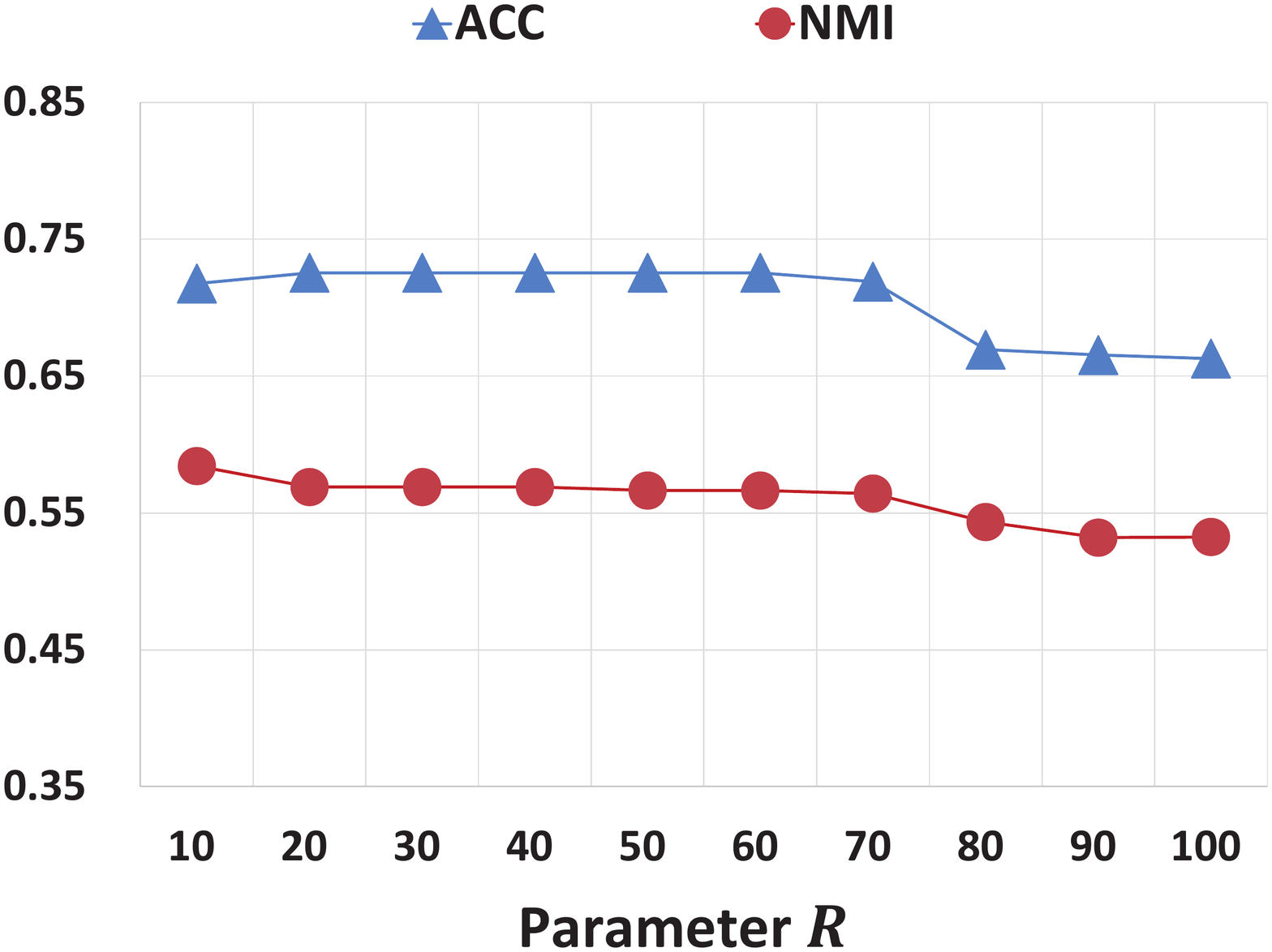}}
  \caption{The performance of RPC with different $R$ on real world uncertain datasets.}
  \label{figure3}
\end{figure}

(1) For parameter $k$, we follow the common practice to set $k$ to the true number of classes in the datasets. (2) For parameter $M$, the investigation results in previous possible world based methods show that setting $M=100$ is enough to obtain satisfactory results \cite{DBLP:conf/icde/VolkRHHL09,DBLP:conf/kdd/ZufleESMZR14}, so we set $M=100$. (3) For parameter $R$, Figure 2 shows the performance of RPC with different $R$ on real world uncertain datasets. From the results, it can be seen that when $R$ is within 10$\sim$70, the clustering performance is always good and stable. When the parameter $R$ is larger than 70, the clustering performance will be affected seriously, which is because that the remaining 30 possible worlds contain many marginal ones. As selecting too many representative possible worlds will result in a waste of time to some extent, in this paper we set $R=10$ and report the corresponding results.

\subsection{Clustering Results}

\textbf{Effectiveness.} From the effectiveness results in Table 2, it can be seen that RPC performs the best. RP-SC and RP-REP respectively perform better than SC and REP, but not as well as RPC. This is because that compared with SC and REP, RP-SC and RP-REP select the representative possible worlds, thus avoiding the negative effects caused by marginal possible worlds. However, compared with RPC, RP-SC and RP-REP do not make use of the consistency principle among different possible worlds. UK-means, CK-means, UK-medoids, MMVar and UCPC perform worse than RPC. The reason is that these algorithms reduce complex probability distributions to a single probability distribution or a determinate value, which may cause the loss of uncertain information. FDBSCAN, FOPTICS and PDBSCAN also perform worse than RPC. The reason is that they rely on the unreasonable independent distance assumption. All in all, in terms of effectiveness, RPC performs much better than the compared algorithms.

\noindent\textbf{Efficiency.} Due to space limit, we only report the efficiency results (in milliseconds) on real world uncertain datasets. Other datasets have the similar trend. From the results in Figure 3, it can be seen that UK-medoids is the slowest. RPC runs faster than FDBSCAN and FOPTICS, but slower than UK-means, CK-means, MMVar, UCPC and PDBSCAN. Among possible world based algorithms, RP-SC, RP-REP and RPC perform almost identically, and they are slower than SC and REP. The reason is that when selecting representative possible worlds, the computation process of Jensen-Shannon divergence is a little complex.

\begin{figure}[t]
  \centering
  \subfigure[Movement]{
    \includegraphics[height=0.23\columnwidth, width=0.29\columnwidth]{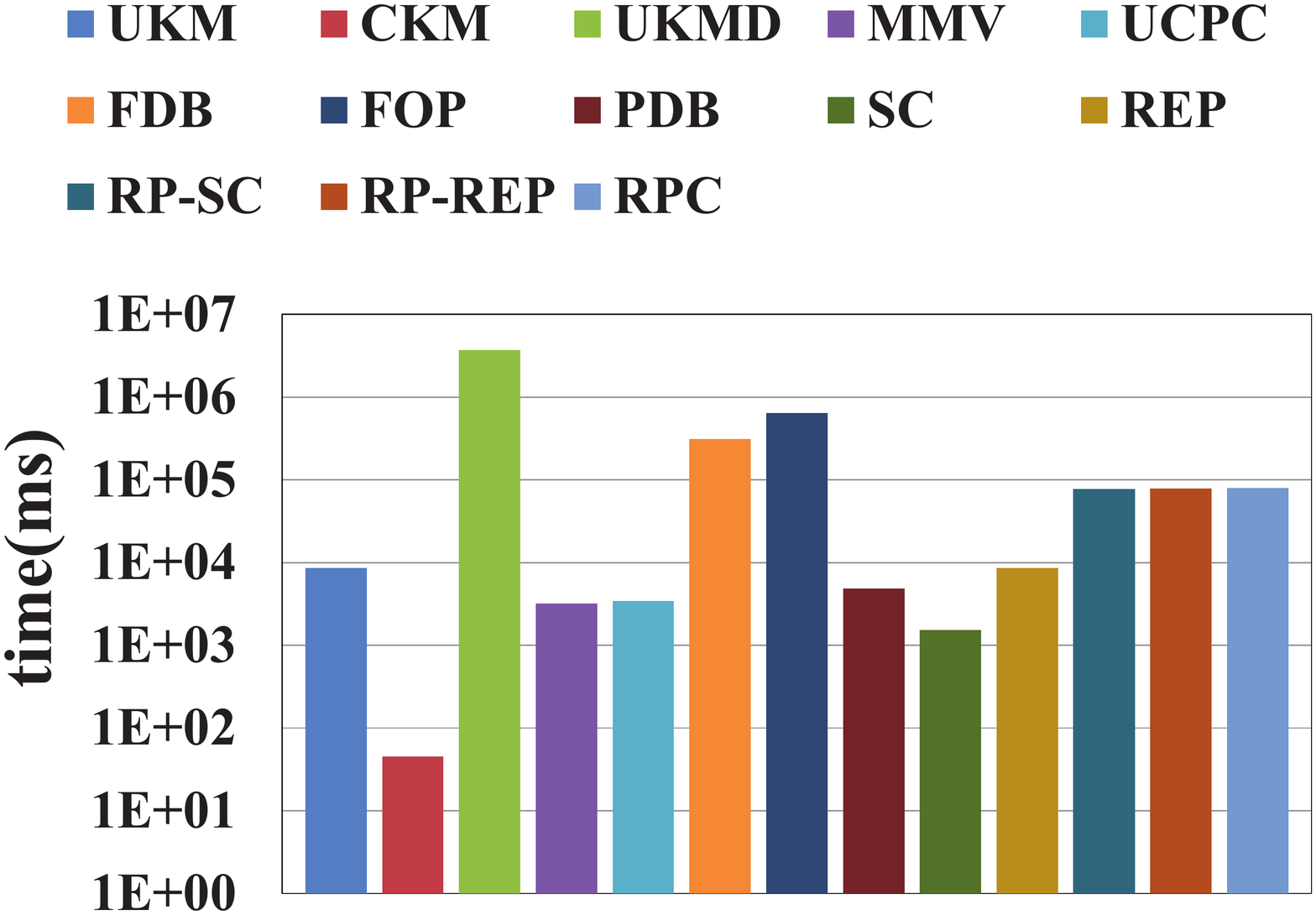}}
  \subfigure[NBA]{
    \includegraphics[height=0.23\columnwidth, width=0.29\columnwidth]{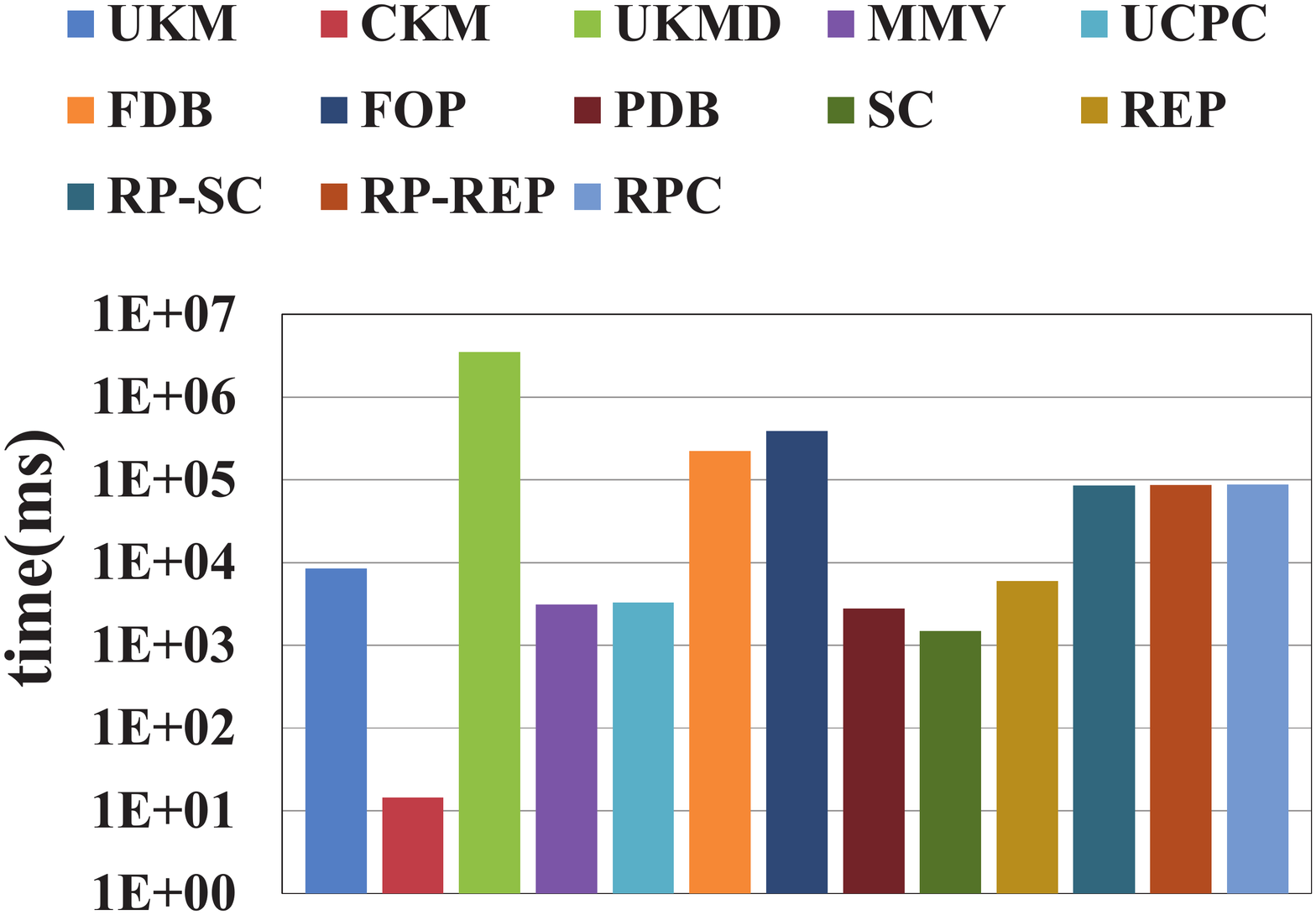}}
  \subfigure[Weather]{
    \includegraphics[height=0.23\columnwidth, width=0.29\columnwidth]{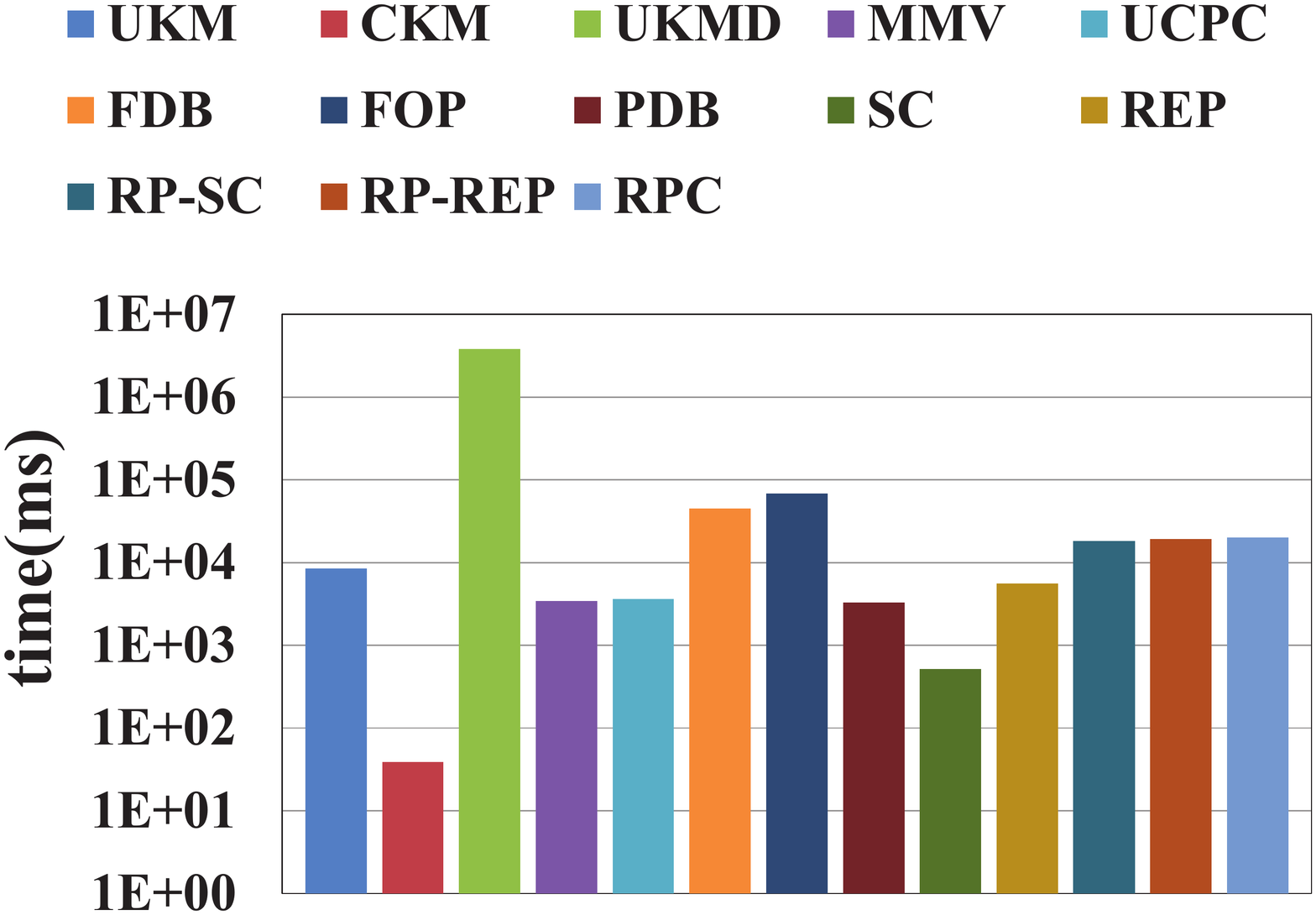}}
  \caption{Clustering results in terms of efficiency.}
  \label{figure4}
\end{figure}

\section{Conclusion}
In this paper, we propose a representative possible world based consistent clustering algorithm for uncertain data. By selecting representative possible worlds, it avoids the negative effects caused by marginal possible worlds. By consistent spectral clustering, it makes use of the consistency principle to achieve better performance. Experimental results show that the proposed algorithm outperforms the state-of-the-art algorithms in effectiveness. For future work, we will extend the idea to uncertain data stream clustering and classification.

\section{Acknowledgments}
This work was supported by National Science Foundation of China (No. 61876028) and the grants 1-ZVJJ and G-YBXV funded by the Hong Kong Polytechnic University.

\bibliographystyle{named}
\bibliography{ijcai19}

\end{document}